\documentclass[journal]{IEEEtran}

\usepackage{siunitx}
\usepackage{hyperref}
\usepackage{pifont}
\usepackage[table]{xcolor}
\usepackage{multirow}
\usepackage{booktabs}
\usepackage{amsmath,amsfonts}
\usepackage{algorithmic}
\usepackage{algorithm}
\usepackage{array}
\usepackage[caption=false,font=normalsize,labelfont=sf,textfont=sf]{subfig}
\usepackage{textcomp}
\usepackage{stfloats}
\usepackage{url}
\usepackage{verbatim}
\usepackage{graphicx}
\usepackage{cite}
\newcommand{\cmark}{\ding{51}}
\newcommand{\xmark}{\ding{55}}

\begin{document}

\title{DreamFormer: Dream Imitation with a Transformer World Model for Language-Conditioned Robotic Manipulation}

\author{Mostafa Kotb$^{1,2,*}$, Cornelius Weber$^{1}$, Muhammad Burhan Hafez$^{3}$, and Stefan Wermter$^{1}$%
\thanks{$^{*}$Corresponding author: Mostafa Kotb}
\thanks{$^{1}$Mostafa Kotb, Cornelius Weber and Stefan Wermter are with the Knowledge Technology Group, Department of Informatics, University of Hamburg, 20146 Hamburg, Germany
        {\tt\footnotesize \{mostafa.kotb, cornelius.weber, stefan.wermter\}@uni-hamburg.de}}%
\thanks{$^{2} $Mostafa Kotb is also with Mathematics Department, Faculty of Science, Aswan University, 81528 Aswan, Egypt
        {\tt\footnotesize m.kotb@sci.aswu.edu.eg}}%
\thanks{$^{3} $Muhammad Burhan Hafez is with School of Electronics and Computer Science, University of Southampton, Southampton SO17 1BJ, UK
        {\tt\footnotesize burhan.hafez@soton.ac.uk}}%
}


\maketitle

\begin{abstract}
We introduce \textit{DreamFormer}, a model-based agent that acquires language-conditioned, multi-task skills by imitating expert demonstrations within the latent imagination of a learned world model. \textit{DreamFormer} first learns a task-agnostic Transformer world model from unstructured play data, then acquires task-specific behaviors by optimizing an intrinsic reward that aligns agent-generated rollouts with expert demonstrations in latent space. Since the policy is trained on-policy inside imagination, it is exposed to its own errors during training, mitigating the covariate shift inherent to offline behavioral cloning. To make long-horizon imagination affordable, \textit{DreamFormer} encodes a high-resolution multi-view robotic observation into a single input token, avoiding both spatial downsampling and the multi-token representations used by prior Transformer world models. On the long-horizon CALVIN benchmark, \textit{DreamFormer} outperforms LUMOS, the comparable model-based agent, on single-environment evaluation (2.52 vs 2.34 average tasks completed per chain of five) while keeping imagination rollouts tractable. Against HULC, the behavior cloning baseline, it nearly doubles performance on zero-shot transfer to an unseen environment (1.30 vs 0.67), indicating that dynamics learned by the world model transfer more readily than a directly cloned policy. This is consistent with the role attributed to internal models in biological agents, where a model of environment dynamics supports behavior in situations not previously encountered. 

\end{abstract}


\section{Introduction}
\IEEEPARstart{A}  foundational hallmark of human and animal intelligence is the ability to
construct rich internal representations of external reality---referred to as \textit{world
models}---that enable biological agents to mentally simulate alternative scenarios, predict
future dynamics, and formulate optimal decisions prior to physical execution \cite{craik}.
Inspired by this mechanism, world models \cite{wm} have emerged as a promising
framework for enhancing the decision-making and predictive capabilities of embodied agents, offering a pathway toward more general and robust intelligent behavior \cite{path}. Such models are used either to learn behavior inside their own imagination \cite{dreamerv2, dreamerv3} or to plan at runtime \cite{tdmpc, qttdm}. Despite recent advances, learning world models that transfer across diverse downstream tasks remains an open challenge.

The difficulty in learning broad-purpose world models stems primarily from two methodological bottlenecks. First, in \textit{online learning} paradigms, the model is continuously updated using data collected by an active policy. This creates a closed-loop exploration bias that inextricably entangles the learned transition dynamics with the specific task being solved. Second, \textit{joint reward modeling} forces the dynamics model to learn representations optimized for a singular, task-specific reward signal. This tightly couples the latent space to one objective, restricting the model's utility for downstream policy learning or real-time planning. Consequently, these paradigms inherently yield \textit{task-specific} world models, significantly hindering knowledge transfer across diverse tasks.

 We argue that constructing broad-purpose world models necessitates a shift toward \textit{offline learning} from diverse offline datasets and \textit{task-agnostic} (i.e., reward-free) training \cite{reward-free}. Once established, these models can serve as versatile foundations, used either to learn downstream task-specific behaviors or to optimize plans to reach a specified goal at runtime. 

Motivated by this principle, we introduce \textit{DreamFormer}, a model-based agent that first
learns a \textit{task-agnostic} world model from unstructured play data, and subsequently
acquires \textit{task-specific} behaviors entirely in the imagination of the
learned latent space. \textit{DreamFormer} builds upon the \textit{Dream Imitation} paradigm
(Section~\ref{sec:2A}) introduced in DITTO \cite{ditto}, which defines an intrinsic reward
encouraging the alignment between imagined policy rollouts and expert trajectories directly in the
model's latent space; optimizing this reward by reinforcement learning induces imitation without
requiring access to expert actions. DITTO, however, validates the paradigm only in single-task settings, training a separate world model for each task without evaluating transfer across tasks. LUMOS
\cite{lumos} extends it to language-conditioned multi-task robotic manipulation, but adopts the DreamerV2 \cite{dreamerv2} world model in its standard configuration, which was developed for $64\times64$ resolution inputs and therefore requires aggressive spatial downsampling of high-resolution robotic observations.

At the architectural level, \textit{DreamFormer} employs an autoregressive, GPT-style Transformer \cite{gpt}, motivated by the proven superiority of Transformers over traditional recurrent dynamics models \cite{qttdm}. However, a critical bottleneck in deploying these architectures for visual control is tokenizing image observations. Conventional tokenizers, such as ViT \cite{vit} or VQ-VAE \cite{discreteAE}, encode a single image into tens or hundreds of tokens. Over extended imagined rollouts, this growth in sequence length inflates both the memory footprint of the key--value cache and the cost of self-attention, which scales quadratically in the number of tokens attended over. 

To mitigate this, prior approaches \cite{storm,twm} have utilized a categorical variational autoencoder (VAE) \cite{dreamerv2} to encode each image into a single discrete token. However, this configuration was developed for low-resolution inputs such as Atari and does not extend to high-resolution
robotics without downsampling \cite{lumos}. \textit{DreamFormer} alleviates this constraint by scaling the categorical VAE: replacing the shallow convolutional encoder with a ResNet-18 backbone and expanding the capacity of the categorical latent space. It encodes full-resolution multi-view observations into a single input token, so that sequence length grows only with the imagination horizon and is independent of input resolution, keeping the memory and attention cost of imagination rollouts bounded by the horizon alone.

The primary contributions of this work are as follows:
\begin{enumerate}
\item \textbf{A Broad-Purpose World Modeling Framework:} We introduce \textit{DreamFormer}, a model-based robotic agent that leverages offline datasets to learn a task-agnostic Transformer world model. By decoupling dynamics learning from singular reward objectives, the model builds versatile representations suitable for transfer across downstream tasks.
\item \textbf{Low-Cost Long-Horizon Rollouts:} We scale the categorical VAE with a ResNet-18 backbone and an expanded latent space, encoding full-resolution multi-view observations at one Transformer token per timestep.  The sequence length of an imagined rollout therefore depends only on the horizon, bounding both the key--value cache and the quadratic cost of self-attention.
\item \textbf{Language-Conditioned Multi-Task Transfer:} We extend the \textit{Dream Imitation} paradigm to train a language-conditioned multi-task policy entirely within the model's latent space. On the CALVIN benchmark, \textit{DreamFormer} nearly doubles the performance of the model-free baseline on zero-shot transfer to an unseen environment and outperforms the model-based baseline on single-environment evaluation.

\end{enumerate}
\section{Background}

\subsection{Dream Imitation}
\label{sec:2A}
Dream Imitation is an imitation learning algorithm proposed in DITTO \cite{ditto} that leverages learned world models to reformulate offline imitation learning as online reinforcement learning within the world model latent space. First, the expert trajectories are mapped into the model latent space; then, an imitation policy is unrolled starting from an arbitrary expert state in the model latent space. Finally, an intrinsic reward $r_t^{int}$ is defined to encourage the alignment of the unrolled policy latent states $s_t^{\pi}$ with expert latent states $s_t^E$ across multiple time steps:
\begin{equation}
r_t^{int}(s_t^E, s_t^\pi) = 
\frac{s_t^E \cdot s_t^\pi}{\max(\|s_t^E\|, \|s_t^\pi\|)^2} .\label{eq:1}
\end{equation}

This specific intrinsic formulation—a modified dot product—was empirically shown to outperform standard L2 and cosine similarity metrics by encouraging similarity without requiring exact matches \cite{ditto}. Furthermore, unlike standard behavioral cloning, optimizing this intrinsic reward via reinforcement learning allows the agent to actively explore alternative trajectories within the simulated environment. By learning the policy 'online' inside the world model's imagination, Dream Imitation effectively mitigates the compounding errors and covariate shift typically associated with offline methods \cite{covariat-shift}. DITTO validates this paradigm in single-task settings, training a separate recurrent world model for each Atari game. \textit{DreamFormer} extends it to language-conditioned robotic manipulation, training a task-agnostic Transformer world model.

\subsection{Unstructured Play Data and Latent Planning}
\label{sec:2B}
Learning from Play (LfP) \cite{playdata} leverages \textit{unstructured, exploratory} interaction data—often collected as humans freely manipulate the environment without predefined goals. Such data is diverse and unlabeled, capturing a wide range of behaviors. However, imitation learning from unstructured play data faces a \textit{multimodality} problem \cite{playdata}, where multiple valid high-level behaviors can connect the same current state $s_c$ and goal state $s_g$. To address this challenge, Play-LMP \cite{playdata} introduces a planning module that encodes the diverse behaviors linking the same $(s_c, s_g)$ pair into a unified latent representation within a latent plan space. The planning module comprises two stochastic encoders: (1) \textbf{a plan recognizer}, which maps an entire sequence into a latent plan, and (2) \textbf{a plan proposer}, which infers a latent plan conditioned on the current and goal states. During inference, the policy is conditioned jointly on the current state, the goal state, and the inferred latent plan to resolve behavioral ambiguity.
\section{Related Work}
\subsection{Transformer World Models}
\label{sec:3A}
Model-based agents aim to learn a world model \cite{wm} that serves as an internal simulation of the environment's dynamics. Then, they leverage the learned model to either train a policy inside the model's imaginations \cite{dreamerv3} or perform real-time planning with Model Predictive Control (MPC) \cite{tdmpc}. Recently, with the success of the Transformer architecture \cite{gpt} in sequence modeling across various domains \cite{transformer-xl, vit}, attention has increased on using Transformers as world models.

One of the earliest attempts in this direction is TransDreamer \cite{transdreamer}, which replaces the Recurrent State-Space Model (RSSM) with a Transformer-based State-Space Model (TSSM), achieving comparable performance to DreamerV2 \cite{dreamerv2}. IRIS \cite{iris} employs VQ-VAE \cite{discreteAE} to encode visual observations into 16 discrete tokens per frame, utilizing an autoregressive Transformer \cite{gpt} for sequence modeling. However, because self-attention scales quadratically with sequence length, applying it over a large number of input tokens leads to severe computational overhead and highly inefficient imagined rollouts during policy training. To achieve more efficient imagination, TWM \cite{twm} leverages a categorical VAE \cite{dreamerv2} to encode visual observations into a single discrete token and employs Transformer-XL \cite{transformer-xl} as its sequence model. However, TWM still applies self-attention across a concatenated sequence of observation, action, and reward tokens. To further improve imagination efficiency, STORM \cite{storm} removes the reward token entirely and fuses the observation and action into a single unified token via a multilayer perceptron (MLP). STORM encodes visual observations into a single discrete token similar to TWM and uses an autoregressive Transformer similar to IRIS. In a parallel effort to stabilize long-horizon rollouts, TWISTER \cite{twister} follows STORM's architecture and supplements the training objective with Action-Conditioned Contrastive Predictive Coding (AC-CPC) to learn robust temporal representations directly in the latent space.

Crucially, however, all the aforementioned Transformer-based agents suffer from two primary limitations. First, they predominantly rely on training task-specific world models (i.e., incorporating explicit reward modeling during training), which entangles the learned latent space with a single objective and restricts downstream policy learning strictly to that task. Second, while most of these methods (with the exception of IRIS) adopt a categorical VAE to encode visual observations into a single discrete token, this configuration was developed for low-resolution, single-view domains such as Atari games \cite{ataribench}. \textit{DreamFormer} resolves both limitations by combining a task-agnostic world model with a scaled categorical VAE designed for high-resolution, multi-view robotics.

\subsection{Language-Conditioned Robotic Manipulation}
\label{sec:3B}
A key capability of a versatile, broad-purpose agent is the ability to execute diverse tasks specified through natural language, providing an intuitive medium for human-robot communication. Consequently, enabling robots to perform language-conditioned, vision-based multi-task learning has become a critical focus. One prominent approach is MCIL \cite{mcil}, which utilizes a single behavioral cloning (BC) policy conditioned on multi-context goals (e.g., a goal image or a natural language instruction). This conditioning enables MCIL to effectively leverage unstructured robot play data \cite{playdata}, despite the fact that only about 1\% of the trajectories are explicitly annotated with language. To alleviate the inherent \textit{multimodality} problem within unstructured play data, MCIL adopts the latent planning module (Section~\ref{sec:2B}) originally introduced in Play-LMP \cite{playdata}. Motivated by these prior studies \cite{mcil, playdata}, the CALVIN benchmark \cite{calvin} provides a formal evaluation framework for language-conditioned, long-horizon, multi-task robotic manipulation. It provides a highly challenging offline dataset comprising $\sim$24 hours of teleoperated, unstructured play data, with approximately 20K language-annotated trajectories. Building upon the foundation of MCIL, HULC \cite{hulc} establishes a robust baseline on CALVIN by integrating several key innovations into a unified framework. Notable features include the use of contrastive learning to better ground language instructions in robotic actions, and the adoption of discrete categorical latent plans rather than continuous ones. Furthermore, HULC introduces a hierarchical control architecture in which a local policy, operating exclusively on gripper-centric observations, is conditioned on global latent plans generated by the planning module. 

More recently, several advanced architectures have significantly pushed the state-of-the-art on the CALVIN benchmark. RoboFlamingo \cite{flamingo} adapts large, pre-trained Vision-Language Models (VLMs) by attaching an explicit policy head to a frozen OpenFlamingo  backbone \cite{openflamingo}. GR-1 \cite{gr1} leverages large-scale video generative pre-training to predict future frames and actions end-to-end via a GPT-style Transformer. 

Most importantly, all of these state-of-the-art approaches fundamentally rely on model-free behavioral cloning paradigms. To the best of our knowledge, LUMOS \cite{lumos} is the only model-based baseline in this domain, applying the \textit{Dream Imitation} \cite{ditto} framework with a DreamerV2 \cite{dreamerv2} world model and downsamples CALVIN's high-resolution multi-view observations to $64\times64$ before encoding.

\subsection{Image Tokenization in Transformer World Models}
\label{sec:3C}
Transformer world models simulate dynamics over a sequence of latent tokens, and the cost of
self-attention scales quadratically in the length of that sequence. The image tokenization
strategy therefore governs both the memory footprint and the attention cost of imagination
rollouts.

Conventional tokenizers such as Vision Transformers (ViT) \cite{vit} or VQ-VAE \cite{discreteAE} successfully preserve spatial resolution by dividing an image into a grid of tokens (e.g., 16 or more tokens per frame), they introduce a severe sequence-length bottleneck. Over extended imagined rollouts, allocating multiple tokens per frame inflates both the key--value cache and
the attention cost, as in architectures such as IRIS \cite{iris}.

Recent efforts have explored various strategies to mitigate this multi-token bottleneck. For instance, CompACT \cite{compact} resamples frozen DINOv3 \cite{dinov3} features into 8 discrete tokens through a set of learnable queries and finite scalar quantization (FSQ) \cite{quantization}. From a different angle, DINO-WM \cite{dino-wm} utilizes frozen DINOv2 \cite{dinov2} features and retains the full set of ViT patch tokens. Instead of reducing the token count, it adopts a ViT-based sequence model with a frame-level causal attention mechanism, enabling the prediction of all per-frame tokens in parallel rather than generating them autoregressively token-by-token. While these are complementary optimizations—one reducing tokens per frame, the other reducing decoding steps per frame—both still allocate multiple tokens per observation. Consequently, the memory footprint and sequence length remain heavily inflated, scaling proportionally with the rollout horizon multiplied by the per-frame token count.

To bound sequence length more aggressively, another line of work \cite{storm, twm, twister} employs a categorical VAE \cite{dreamerv2} to encode observations into a single discrete token. However, standard categorical VAEs are architecturally constrained: they rely on shallow CNN encoders with a categorical latent space of 32 categories (each containing 32 classes), and are heavily optimized to operate exclusively on low-resolution images. 
 
\section{Methodology}
We introduce \textit{DreamFormer}, a model-based robotic framework designed to acquire language-conditioned, multi-task  policy entirely within the latent imagination of a learned world model. The training comprises two stages: First, the world model is trained in \textit{a task-agnostic} fashion using an unstructured play dataset. Second, an actor-critic agent is trained inside the world model's imagined latent space via \textit{Dream Imitation} \cite{ditto} (see Section~\ref{sec:2A}) to synthesize a language-conditioned, multi-task policy. This section is organized as follows: Section~\ref{sec:4A} formalizes the problem. Section~\ref{sec:4B} presents our proposed scaling mechanism for categorical VAE. In Section~\ref{sec:4C} details the architecture and training objective of the Transformer world model. Section~\ref{sec:4D} describes the architecture and the training of the agent. Finally, Section~\ref{sec:4E} provides the agent real-time inference.

\subsection{Problem Formulation}
\label{sec:4A}
We perform goal-conditioned imitation learning using a large, unlabeled, and unstructured fixed play dataset $\mathcal{D} = \{\tau_i\}_{i=1}^{N}$, where each trajectory $\tau_i = \{(s_0, a_0), \ldots, (s_k, a_k)\}$ represents a long, unsegmented sequence of state–action pairs. Here, $s \in \mathcal{S}$ denotes the state–observation space, and $a \in \mathcal{A}$ represents the continuous action space. The dataset is subsequently segmented and relabeled, producing $\mathcal{D}_{play} = \{(\tau, s_g)_i\}_{i=0}^{|\mathcal{D}_{play}|}$, which pairs each goal state $s_g$ with its corresponding demonstration trajectory $\tau$. Following the multi-context imitation learning framework proposed in MCIL \cite{mcil}, goals can be expressed either as goal states $s_g$ or free-form language instructions $l$. Consequently, a small subset of $\mathcal{D}_{play}$ is annotated with hindsight language instructions, resulting in $\mathcal{D}_{(play,lang)} = \{(\tau, l)_i\}_{i=0}^{|\mathcal{D}_{(play,lang)}|}$. We aim to learn a goal-conditioned policy $\pi_{\theta}(a_t|s_t,g)$, where $g\in\mathcal{G}$ represents a goal state $s_g$ or a language instruction $l$.

\subsection{Scaling Categorical VAE}
\label{sec:4B}

The standard categorical VAE, initially introduced in DreamerV2 \cite{dreamerv2}, encodes visual observation $o_t$ into a latent categorical distribution comprising 32 categories, each containing 32 classes ($32\times32$). To construct a stochastic latent state $z_t$, a single class is sampled from each category, yielding 32 one-hot vectors, each of length 32. Flattening $z_t$ yields a sparse binary vector of dimension 1024, containing exactly 32 active bits (ones) while the remaining elements are zeros. This standard ($32 \times 32$) categorical VAE is heavily optimized to operate on $64 \times 64$ low-resolution, single-view observations, such as those found in Atari environments. To operate on high-resolution, multi-view robotic environments, it requires downsampling to meet the $64 \times 64$ input constraint, as seen in LUMOS \cite{lumos}. 

We propose to scale the categorical VAE to directly operate on high-resolution observations without  downsampling. First, to extract rich, fine-grained representations, we replace the conventional shallow CNN encoder with an expressive, trainable ResNet-18 backbone, further enhanced with a spatial softmax layer \cite{spatialsoftmax}. The spatial softmax layer reduces each feature map to its spatial expectation, explicitly encouraging the extraction of localized, keypoint-like features. This inductive bias is particularly suited for robotic manipulation, where precise object and gripper positions matter more than global visual textures. Second, we expand the capacity of the categorical latent space to 64 categories, each with 32 classes ($64\times32$). The flattened stochastic latent state $z_t$ expands to a 2048-dimensional vector with 64 active bits. As a consequence of expanding the categorical capacity, we observed training instability: early in training the categorical posterior becomes near-deterministic, concentrating probability mass on a single class per category and leaving much of the expanded latent space unused. To stabilize training, we introduce an entropy regularizer. In the next section, we explain in detail the overall training objective proposed to effectively train our scaled categorical VAE alongside the Transformer dynamics model.
\begin{figure}[!t]
    \centering
    \includegraphics[width=0.5\textwidth]{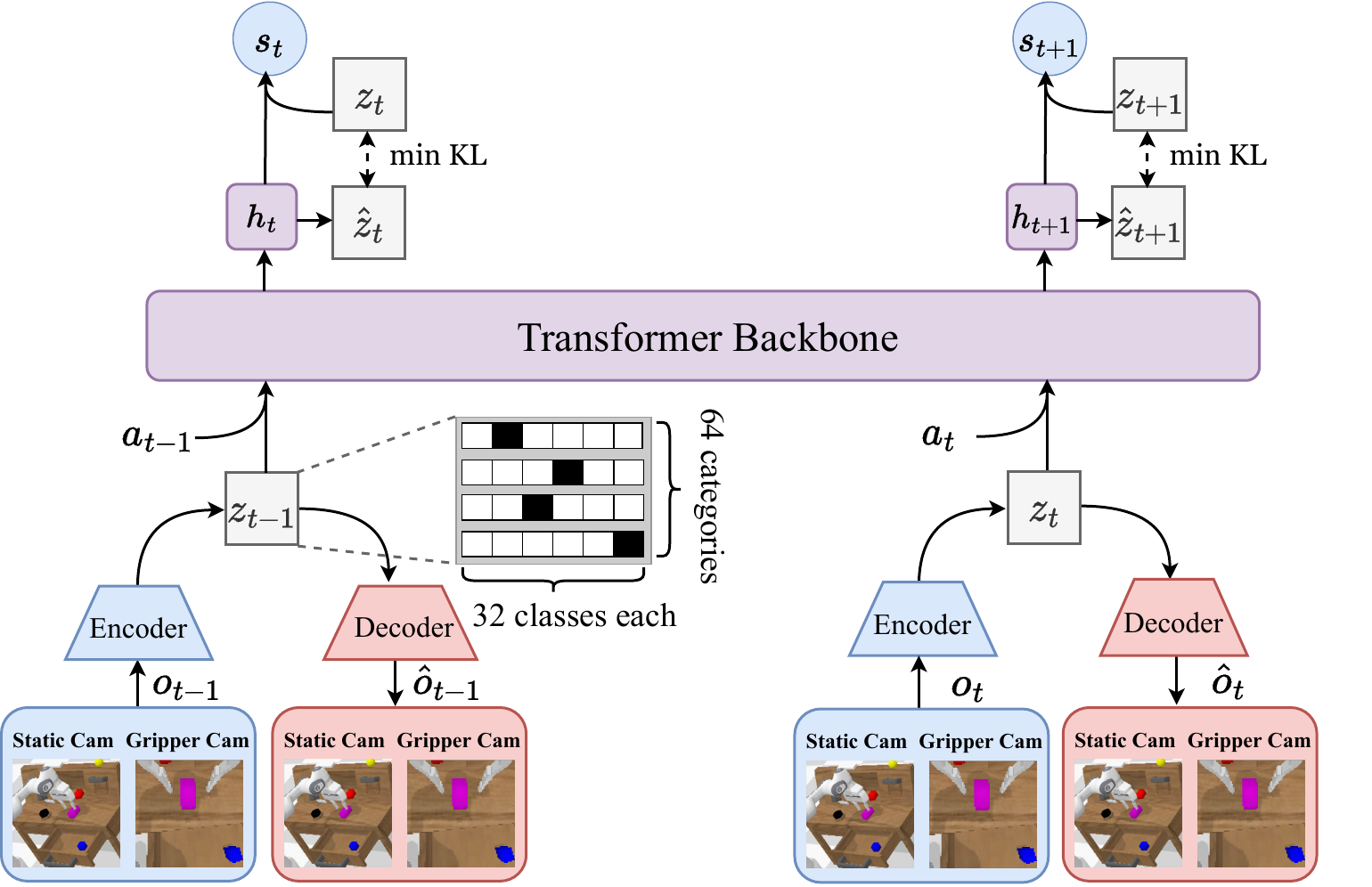}
    \caption{\textbf{Transformer world model} is trained by reconstructing the observation $o_t$ and minimizing the KL divergence between the predicted prior distribution $\hat{z}_t$ and the posterior distribution $z_t$. To accommodate high-resolution, multi-view observations, a trainable ResNet-18 is employed as the visual encoder, and the capacity of the categorical latent space is doubled to 64 categories, each with 32 classes $(64\times32)$.}
    \label{fig:1}
\end{figure}
\subsection{World Model Learning}
\label{sec:4C}
The first training stage involves learning a task-agnostic, latent world model that captures the environment dynamics within a compact latent representation, effectively serving as an internal simulation of the real environment. This internal simulation enables sample-efficient behavior learning, allowing the agent to be trained within the learned model without requiring direct interaction with the environment. Since learning an accurate world model is crucial for effective downstream behavior learning, we employ a Transformer world model as illustrated in Fig.~\ref{fig:1}.

The sequence model is implemented as a GPT-like Transformer \cite{gpt} with masked self-attention to predict the next stochastic state $\hat{z}_{t+1}$ conditioned on the previous states $z_{1:t}$ and actions $a_{1:t}$. The Transformer outputs hidden states $h_t$, which are concatenated with the corresponding stochastic states $z_t$ to form the model state $s_t = \{h_t, z_t\}$. We utilize the state-action fusion strategy proposed in STORM \cite{storm}, wherein the latent stochastic states $z_t$ and actions $a_t$ are concatenated and passed through a multilayer perceptron (MLP), yielding a single token per time step $t$. The world model components are the following:
\begin{equation}
\setlength{\arraycolsep}{6pt} 
\begin{array}{@{}l@{\quad}r@{\;}c@{\;}l@{}}
\textit{Encoder network:}   & z_t        & \sim & q_{\phi}(z_t \mid o_t) \\
\textit{Transformer Backbone:} & h_t        & =    & f_{\phi}(z_{1:t-1}, a_{1:t-1}) \\
\textit{Dynamics predictor:}& \hat{z}_t  & \sim & p_{\phi}(\hat{z}_t \mid h_t) \\
\textit{Decoder network:}   & \hat{o}_t  & \sim & p_{\phi}(\hat{o}_t \mid z_t) 
\end{array}
\label{eq:2}
\end{equation}

Our world model is trained in a task-agnostic fashion, therefore \textit{reward predictor} component is not required. The expanded ($64\times32$) latent is more prone to the collapse described in Section~\ref{sec:4B}: with twice as many categorical variables, there is more capacity available to go unused, and we observe the posterior concentrating on a single class per category early in training. To counteract this, we add a regularizer that maximizes the normalized entropy of the categorical posterior. Ultimately, the world model is trained end-to-end, optimizing the following total objective:
\begin{equation}
\begin{split}
\mathcal{L}(\phi) &= \frac{1}{BT}\sum_{b=1}^{B}\sum_{t=1}^{T} \Big[ \beta_1\mathcal{L}_t^{\text{rec}}(\phi) - \beta_2\mathcal{L}_t^{\text{ent}}(\phi) \\
&\quad + \beta_3\mathcal{L}_t^{\text{rep}}(\phi) + \beta_4\mathcal{L}_t^{\text{dyn}}(\phi) \Big],
\end{split}
\label{eq:3}
\end{equation}
where $B$ is the batch size and $T$ is the sequence length. $\mathcal{L}_t^{\text{rec}}$ represents the reconstruction loss, computed as the mean squared error (MSE) and weighted by $\beta_1$. $\mathcal{L}_t^{\text{ent}}$ represents the entropy regularizer weighted by $\beta_2$, and computed as follows:
\begin{equation}
\mathcal{L}_{\text{ent}} = \frac{\mathcal{H}(z_t)}{\mathcal{H}_{\max}},
\label{eq:4}
\end{equation}
where the absolute entropy $\mathcal{H}(z_t)$ is normalized by the maximum entropy $\mathcal{H}_{\max} = N \log K$, given $N=64$ categories and $K=32$ classes. The absolute entropy of the latent state $z_t$ is defined as:
\begin{equation}
\mathcal{H}(z_t) = -\sum_{i=1}^{N} \sum_{j=1}^{K} p_{i,j} \log p_{i,j},
\label{eq:5}
\end{equation}
where $p_{i,j}$ denote the predicted posterior probability of the $j$-th class within the $i$-th category. Both $\mathcal{L}_t^{\text{dyn}}$ and $\mathcal{L}_t^{\text{rep}}$ losses are expressed as Kullback–Leibler (KL) divergence but differ in weighting and the use of the stop-gradients $sg(.)$ operation:
\begin{subequations}
\begin{align}
\mathcal{L}^{\text{dyn}}_{t}(\phi) &= \max\left(1, 
\mathrm{KL}\left[
sg\big(q_{\phi}(z_{t} \mid o_{t})\big) \,\Big\|\, p_{\phi}(\hat{z}_{t} \mid h_t)
\right]\right), \\[1pt]
\mathcal{L}^{\text{rep}}_{t}(\phi) &= \max\left(1, 
\mathrm{KL}\left[
q_{\phi}(z_{t} \mid o_{t}) \,\Big\|\, sg\big(p_{\phi}(\hat{z}_{t} \mid h_t)\big)
\right]\right).
\end{align}
\label{eq:6}
\end{subequations}

Dreamer (V2 and V3) \cite{dreamerv2, dreamerv3} balances the two KL terms with fixed coefficients tuned for a $(32\times32)$ latent space. In our $(64\times32)$ space the KL is summed over twice as many categorical variables, so the same coefficients impose a proportionally stronger penalty on the posterior and contribute to the collapse described above. We instead ramp both coefficients over the first 250K steps with a cosine schedule: $\beta_3$ from 0.1 to 0.5 and $\beta_4$ from 1.0 to 2.0. Early training is thus weighted an order of magnitude toward the dynamics term, which constrains the transition model to track the posterior while leaving the encoder comparatively free to fit high-resolution observations; the schedule then settles at a 4:1 ratio, matching Dreamer's balance at a larger overall scale. Furthermore, to prevent the high-capacity ResNet-18 encoder from memorizing scene appearance and encourage it to learn robust features, we apply random pixel-shift augmentation \cite{augmentation}. Crucially, a single shift is sampled per trajectory and applied uniformly to every frame, rather than independently per frame: this makes the encoder robust to global camera displacement while leaving inter-frame motion (i.e., preserving relative temporal dynamics). Finally, sampling $z_t$ from a categorical distribution is not differentiable, so we employ straight-through gradients \cite{straight-through}.

\subsection{Agent Behavior Learning}
\begin{figure*}[!t]
    \centering
    \includegraphics[width=0.99\textwidth]{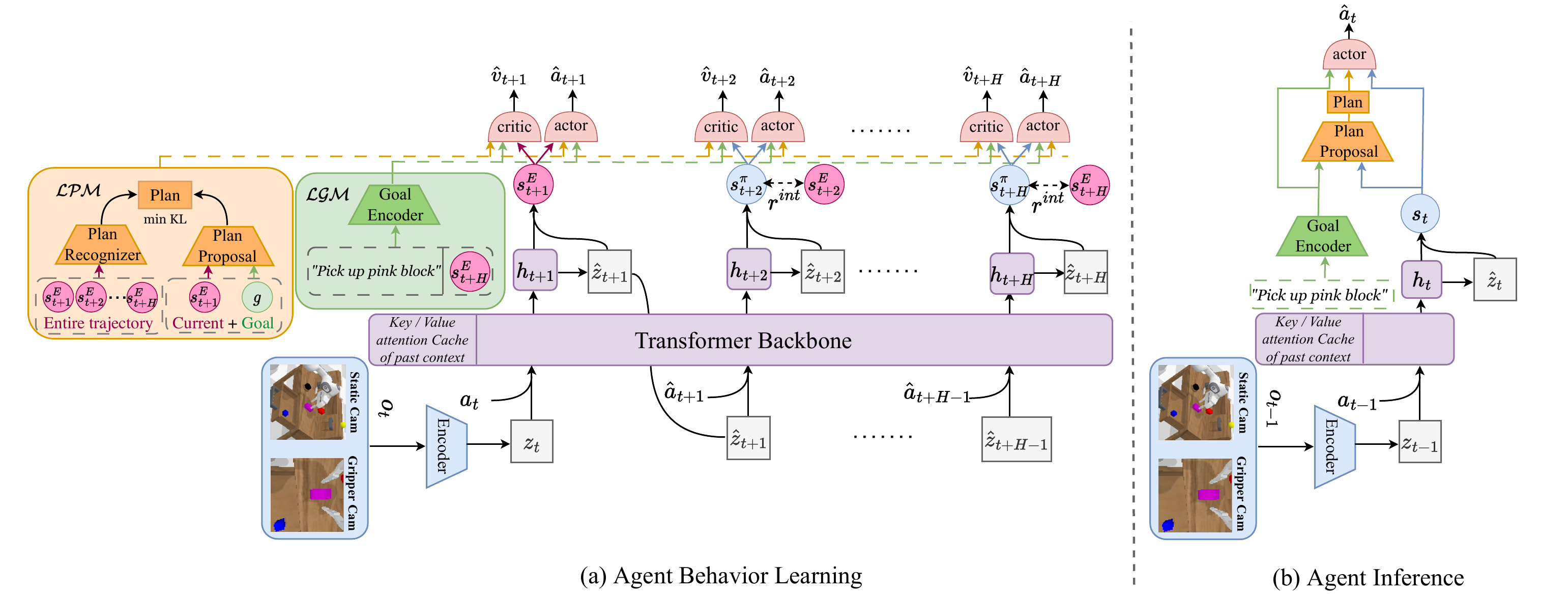}
    \caption{\textbf{DreamFormer} learns goal-conditioned behaviors within the latent space of a learned world model. (a) During training, imagined trajectories are unrolled from an expert state, initialized with a key--value cache over a context window of preceding steps. Goals are encoded by the latent goal module $(\mathcal{LGM})$ from either a language instruction or the final expert latent state, and the latent plan module $(\mathcal{LPM})$ supplies a high-level behavioral prior: the plan is drawn from the plan recognizer, which sees the entire expert trajectory, while the plan proposal is trained to match it from the current state and goal alone. The actor is trained by maximizing an intrinsic reward that rewards alignment between imagined and expert latent trajectories. (b) During inference, the actor acts based on the current latent state $s_t = \{h_t, \hat{z}_t\}$, with $h_t$ computed over a sliding context window of recent real observations, guided by a language command. The latent plan comes from the plan proposal rather than the recognizer, conditioning the actor on the intended behavior.}
    \label{fig:2}
\end{figure*}
\label{sec:4D}
DreamFormer trains an actor-critic agent within the latent space of the learned world by unrolling imagined trajectories starting from an arbitrary expert state and encourages the alignment of the unrolled policy states $s_t^{\pi}$ with the expert states $s_t^E$. Specifically, an intrinsic reward $r^{int}$ is defined by Eq.~\ref{eq:1} to measure the divergence between the agent rollouts and expert demonstrations. The actor–critic agent is trained to maximize this intrinsic reward using standard reinforcement learning. Here, the state $s_t = \{h_t,z_t\}$ denotes the model state, which is composed of the hidden state $h_t$, encoding historical information from the previous time steps $(1:t-1)$, and the stochastic state $z_t$, capturing information from the current time step $t$. Following LUMOS, we learn a goal-conditioned policy $\pi_{\theta}(a_t \mid s_t,g)$ and a value function $v_{\psi}(s_t \mid g )$ from the dataset $\mathcal{D}_{play}$ and $\mathcal{D}_{(play,lang)}$. Fig.~\ref{fig:2}(a) shows the agent behavior learning process in detail.

\textit{Imagination Rollouts:} During policy training, a batch of expert trajectories $\{\tau_i\}_{i=1}^{B}$ is sampled, each of length $H$, where $H$ denotes the imagination horizon. First, these expert trajectories are mapped into the model's latent space to obtain the sequence of expert states $\{s_{t+1}^E, s_{t+2}^E, \ldots, s_{t+H}^E\}$ \textit{(depicted as pink circles)}. Starting from the initial expert state $s_{t+1}^E$, the imagined trajectories are autoregressively unrolled using the policy, the prior distribution, and the Transformer backbone to generate the corresponding policy states $\{s_{t+2}^{\pi}, s_{t+3}^{\pi}, \ldots, s_{t+H}^{\pi}\}$ \textit{(depicted as blue circles)}. Since the unrolling process is explicitly conditioned on the initial expert state, there is no corresponding policy state at timestep $t+1$. The imagination process is initialized with a key--value cache computed over the context window $\mathcal{C}$, and the cache grows with each imagined step, so every predicted state attends over the full context and all preceding imagined states without truncation. Finally, the intrinsic reward $r^{int}$ is computed via Eq.~\ref{eq:1} to measure the alignment between each corresponding state pair $(s_{k}^E, s_{k}^{\pi})$ across $k=t+2,\ldots,t+H$.

\textit{Latent Goal Module:} to learn a goal-conditioned actor–critic agent, we employ a latent goal module $(\mathcal{LGM})$ \textit{(depicted as the green box)} that maps the goal into a latent goal representation $\mathbb{R}^g$, where $g$ denotes the dimensionality of the latent goal space. The goal input can be either a language instruction or a latent state corresponding to the final expert state $s_{t+H}^E$. Similar to HULC \cite{hulc}, separate goal encoders are used for each goal modality. In order to embed raw text into a semantically meaningful vector space, we utilize the pre-trained \textit{paraphrase-MiniLM-L3-v2} model~\cite{bert}.

\textit{Latent Plan Module:} to address the \textit{multimodality} challenge (see Section~\ref{sec:2B}) encountered in learning control from unstructured play data, we employ a latent plan module $(\mathcal{LPM})$\textit{(depicted as the orange box)}. It consists of two stochastic encoders: (1) \textit{a plan recognizer}, which maps the entire latent state trajectory $\{s_{t+1}^E,s_{t+2}^E,\ldots,s_{t+H}^E\}$ into a latent plan, and (2) \textit{a plan proposer}, which proposes a latent plan conditioned on the current latent state $s_{t+1}^E$ and a latent goal $g$. During training, we minimize the KL divergence $\mathcal{L}_{\text{KL}}$ between the two encoders to ensure that the plan proposal accurately captures and reproduces the observed behavior. Similar to HULC, we employ a multimodal transformer encoder to build a contextualized representation of the latent trajectories, mapping them into categorical latent plans of 32 categories, each with 32 classes and optimizing them using straight-through gradients \cite{straight-through}.

\textit{Semantic Alignment:} To enable the agent to follow language instructions, we must address the symbol grounding problem \cite{symbol}—the fundamental challenge of mapping abstract linguistic commands to the robot’s perception and actions. Following the approach in LUMOS, we bridge this modality gap by aligning the world model’s latent representations with the extracted language features. This semantic alignment is achieved via a CLIP-style contrastive loss \cite{clip}, denoted as $\mathcal{L}_{\text{contrast}}$, which leverages in-batch negative sampling to ensure efficient and robust training.

\textit{Actor and Critic models:} Both the actor and critic models are conditioned on the latent state $s_t$ and the latent goal $g$. The actor samples actions aimed at maximizing the intrinsic reward $r^{\text{int}}$, while the critic predicts value estimates for the same latent state–goal pair, representing the expected future discounted rewards. The models are defined as follows:
\begin{equation}
\begin{aligned}
\text{Actor:} \quad 
    & a_t \sim \pi_{\theta}(a_t \mid s_t, g), \\[4pt]
\text{Critic:} \quad 
    & v_{\psi}(s_t, g) \approx 
      \mathbb{E}_{\pi_{\theta}, p_{\phi}} 
      \left[ 
        \sum_{t=0}^{H} \gamma^{t} r_t^{\text{int}} 
      \right],
\end{aligned}
\label{eq:7}
\end{equation}
where $\gamma$ is a discount factor and $H$ is the imagination horizon. We implement the actor and critic models as two MLPs, and the actor outputs a tanh-transformed Gaussian \cite{sac}, facilitating reparameterized sampling for backpropagation. The world model remains fixed during behavior learning to prevent agent gradients from modifying its learned representations. We adopt the training objectives from LUMOS, where the critic objective is to regress the computed value estimate to the $\lambda$-target \cite{sutton}, denoted by $V^{\lambda}$, as follows:
\begin{equation}
\mathcal{L}(\psi) \doteq \mathbb{E}_{\pi_\theta, p_\phi} \left[ 
\sum_{t=1}^{H-1} \frac{1}{2} \left( v_\psi(s_t) - sg(V_t^{\lambda}) \right)^2 
\right],
\label{eq:8}
\end{equation}
where $sg(.)$ is the stop-gradients operation, and a target network is employed to provide target values, whose parameters are an exponential moving average (EMA) of the critic network. The actor objective leverages backpropagation through the learned dynamics to maximize value estimates while minimizing the KL divergence between plan encoders and the contrastive loss aligning latent trajectories with language:
\begin{equation}
\mathcal{L}(\theta) \doteq \mathbb{E}_{\pi_\theta, p_\phi} \left[ 
\sum_{\tau=t}^{t+H}  \left( -V^{\lambda}(s_{\tau}) + \alpha_1\mathcal{L}_{\text{KL}} + \alpha_2\mathcal{L}_{\text{contrast}} \right) 
\right],
\label{eq:9}
\end{equation}
where $\theta$ represents the parameters of the actor, the latent plan module, and the latent goal module.
\subsection{Agent Inference}
\label{sec:4E}
At inference (Fig.~\ref{fig:2}(b)), the policy is conditioned solely on the language instruction, while the proposal encoder infers the latent plan from the current state and latent goal. The model does not imagine at execution time: the key--value cache holds a context window $\mathcal{C}$ of the most recent real observations and slides forward as the episode proceeds, so it provides history to the sequence model at bounded cost.
\section{Experiments and Results}
This section evaluates \textit{DreamFormer} on the CALVIN benchmark \cite{calvin}. We compare against learning-based
comparable methods. We then quantify the contribution of individual components through an ablation study. Finally, we analyse imagination rollouts along two axes: the fidelity of the world model’s open-loop predictions, and the computational
cost of rollouts compared with other Transformer world models.
\begin{figure*}[!t]
    \centering
    \includegraphics[width=0.80\textwidth]{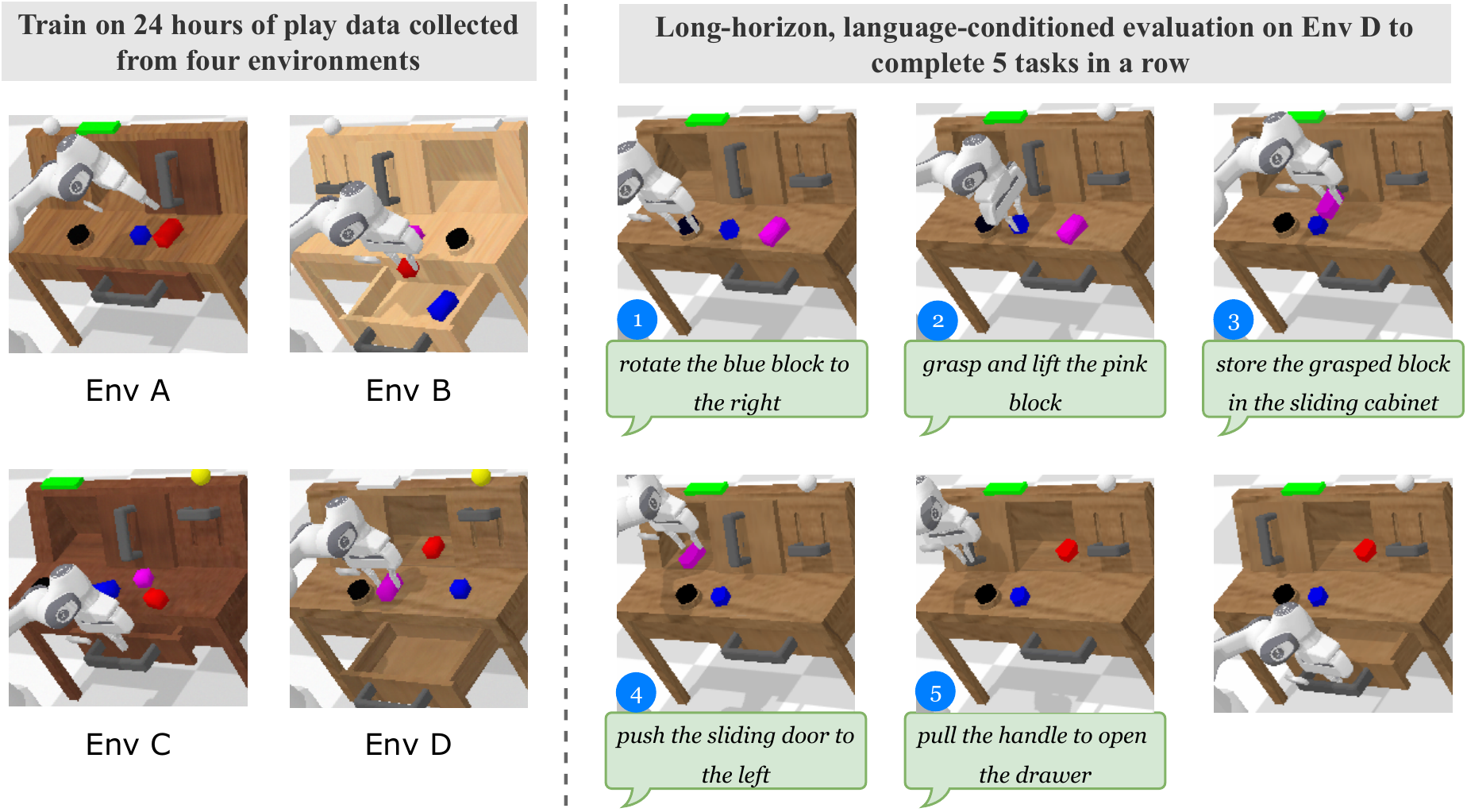}
    \caption{\textbf{CALVIN Benchmark}. (Left) the four environments A, B, C, D. (Right) long-horizon, language-conditioned evaluation with 5 tasks in a row.}
    \label{fig:3}
\end{figure*}
\subsection{Experimental Setup}
\label{sec:5A}
\textit{1) Benchmark:} The CALVIN challenge \cite{calvin} consists of four environments A, B, C, D with similar setup but vary in the desk color and the layouts of items, as shown in Fig.~\ref{fig:3} (Left). Each environment contains a 7-DOF Franka Emika Panda robot arm and a desk with a sliding door, a drawer, an LED controlled by a black button, and a light bulb controlled by a switch. Three colored rectangular blocks are also present. The blocks are  movable objects and can appear at different positions on the tabletop. The challenge offers a dataset of $\sim$24 hours of teleoperated unstructured play data with multiple sensor modalities, in which only $1\%$ is annotated with language instructions. All methods are evaluated on the long-horizon benchmark, which consists of 1000 unique sequences of instruction chains.  The agent's goal is to sequentially solve up to five tasks using only onboard sensors, relying exclusively on language instructions to transition between tasks, as shown in Fig.~\ref{fig:3} (Right). The benchmark offers three settings of increasing difficulty: D→D trains and evaluates in a single environment, ABCD→D trains on all four environments and evaluates on D, and ABC→D trains on A, B and C and evaluates zero-shot on the unseen environment D.
 
\textit{2) Baselines:} We compare against a range of methods, divided into two categories.

\textit{(1) No visual pretraining:} represents comparable methods to our approach where the vision encoder is trained from scratch without any pretraining phase. This category includes:
\begin{itemize}
    \item \textit{LUMOS} \cite{lumos}: The key comparison, and the only model-based baseline that learns a language-conditioned policy inside the imagination of a learned world model via Dream Imitation \cite{ditto}. It uses a recurrent world model and downsamples observations to $64\times64$.

    \item \textit{GCBC} \cite{playdata}: Goal-conditioned behavior cloning trains the policy by maximizing the log likelihood of each action taken at each time step and does not use latent plans.

    \item \textit{MCIL} \cite{mcil}:  Multi-context imitation learning trains a single behavioral cloning policy conditioned on multi-context goals and employs latent plans.

    \item \textit{HULC} \cite{hulc}: Hierarchical Universal Language Conditioned policy, which learns discrete categorical latent plans, aligns video and language representations with a contrastive loss, and executes actions through a local policy operating in the gripper-camera frame.
\end{itemize}

\textit{(2) Pretrained visual/vision-language backbone:} represents SOTA methods that utilize a pretrained vision encoder or employ vision-language backbone. This category includes:
\begin{itemize}
    \item \textit{GR-1} \cite{gr1}: leverages large-scale video generative pretraining to predict future frames and actions end-to-end via a GPT-style Transformer.

     \item \textit{RoboFlamingo} \cite{flamingo}: utilizes OpenFlamingo \cite{openflamingo} vision-language backbone which consists of 3B parameters.

     \item \textit{RT-1} \cite{rt1}: a decoder-only Transformer policy generates actions in an end-to-end manner. It uses an ImageNet-pretrained convolutional neural network (EfficientNet) conditioned on the language instruction via FiLM layers.  

     \item \textit{MT-R3M} as reported in \cite{gr1}: leverages the pretrained vision model R3M \cite{R3M}, and utilizes a GPT-style Transformer to output actions.
\end{itemize}

All methods in both categories, including ours, use a frozen pretrained sentence encoder for language conditioning; the distinction here concerns visual representations only. Category-1 constitutes our controlled comparison: these methods learn visual representations from CALVIN data alone, so differences in performance are attributable to modelling choices rather than to external data.  Category-2 is reported to situate \textit{DreamFormer} within the broader CALVIN literature; these methods inherit visual or vision-language representations trained on corpora orders of magnitude larger than CALVIN, and we therefore treat them as reference points for absolute task performance rather than as matched baselines.
\begin{table*}[!t]
\caption{Performance of all methods on CALVIN D$\rightarrow$D split. Methods are grouped by model class (model-based and model-free). Bold marks the best result within each group. Parentheses give the standard deviation over three seeds}
\label{tab:1}
\centering
\small
\begin{tabular}{lccccccc}
\toprule
\multirow{2}{*}{Method} & Downsample
 & \multicolumn{6}{c}{Number of Instructions in a Row (1000 chains)} \\
\cmidrule(lr){3-8}
 & $64{\times}64$ & 1 & 2 & 3 & 4 & 5 & Avg. Len. \\
\midrule
\rowcolor{gray!40} \multicolumn{8}{c}{\textit{Model-Based Methods}} \\
\midrule
LUMOS \cite{lumos}
& \cmark & 80.7\% (1.0) & 59.3\% (1.4) & 42.6\% (1.8) & 30.7\% (1.2) & 21.1\% (0.8) & 2.34 (0.05) \\
DreamFormer (ours)
& \xmark & \textbf{82.3\%} (1.7) & \textbf{64.1\%} (2.2) & \textbf{47.1\%} (2.8) & \textbf{34.2\%} (2.0) & \textbf{24.3\%} (2.0) & \textbf{2.52} (0.11) \\
\midrule
\rowcolor{gray!40} \multicolumn{8}{c}{\textit{Model-Free Methods}} \\
\midrule
GCBC \cite{playdata}
& \xmark & 64.7\% (4.0) & 28.4\% (6.2) & 12.2\% (4.1) & 4.9\% (2.0) & 1.3\% (0.9) & 1.11 (0.3) \\
MCIL \cite{mcil}
& \xmark & 76.4\% (1.5) & 48.8\% (4.1) & 30.1\% (4.5) & 18.1\% (3.0) & 9.3\% (3.5) & 1.82 (0.2) \\
HULC (Local Policy) \cite{hulc}
& \xmark & \textbf{82.7\%} (0.3) & \textbf{64.9\%} (1.7) & \textbf{50.4\%} (1.5) & \textbf{38.5\%} (1.9) & \textbf{28.3\%} (1.8) & \textbf{2.64} (0.05) \\
HULC (No Local Policy)\cite{hulc}
& \xmark & 78.4\% (1.4) & 56.2\% (2.0) & 40.4\% (2.4) & 29.5\% (1.7) & 20.1\% (1.3) & 2.24 (0.06) \\
\bottomrule
\end{tabular}
\end{table*}
\begin{table*}[!t]
\caption{Performance of all methods on CALVIN ABCD$\rightarrow$D and ABC$\rightarrow$D splits. Methods are grouped by pretraining regime (w/o pretraining and  w/ pretraining). Bold marks the best result within each group. Parentheses give the standard deviation over three seeds; the remaining baselines report single-run values}
\label{tab:2}
\centering
\small
\begin{tabular}{lccccccc}
\toprule
\multirow{2}{*}{Method} & \multirow{2}{*}{Pretraining}
 & \multicolumn{6}{c}{Number of Instructions in a Row (1000 chains)} \\
\cmidrule(lr){3-8}
 &  & 1 & 2 & 3 & 4 & 5 & Avg. Len. \\
\midrule
\rowcolor{gray!40} \multicolumn{8}{c}{\textit{ABCD$\rightarrow$D}} \\
\midrule
MCIL \cite{mcil}   &  \xmark & 37.3\% & 2.7\% & 0.2\% & 0.0\% & 0.0\% & 0.40 \\
HULC \cite{hulc}   &  \xmark & 88.9\%  & 73.3\%  & 58.7\%  &47.5\%  & 38.3\%  & 3.06 (0.07) \\
DreamFormer (ours) &  \xmark & \textbf{89.1\%}  & \textbf{74.0\%}  & \textbf{60.1\%}  & \textbf{49.1\%}  & \textbf{38.4\%}  & \textbf{3.11} (0.05)  \\
\midrule
RT-1 \cite{rt1}    & ImageNet & 84.4\% & 61.7\% & 43.8\% & 32.3\% & 22.7\% & 2.45 \\
MT-R3M \cite{gr1}  & Ego4D (R3M) & 75.2\% & 52.7\% & 37.5\% & 25.8\% & 16.3\% & 2.08 \\
RoboFlamingo \cite{flamingo} & OpenFlamingo 3B & \textbf{96.4\%} & 89.6\% & 82.4\% & 74.0\% & 66.0\% & 4.09 \\
GR-1 \cite{gr1}    &  Video Generative & 94.9\% & \textbf{89.6\%} & \textbf{84.4\%} & \textbf{78.9\%} & \textbf{73.1\%} & \textbf{4.21} \\
\midrule
\rowcolor{gray!40} \multicolumn{8}{c}{\textit{ABC$\rightarrow$D}} \\
\midrule
MCIL \cite{mcil}   &  \xmark &  30.4\% & 1.3\% & 0.2\% & 0.0\% & 0.0\% & 0.31 \\
HULC \cite{hulc}   &  \xmark & 41.8\%  & 16.5\%  & 5.7\%  & 1.9\%  & 1.1\%  & 0.67 (0.10) \\
DreamFormer (ours) &  \xmark & \textbf{62.3\%} & \textbf{35.4\%} & \textbf{18.3\%} & \textbf{9.4\%} & \textbf{4.5\%} & \textbf{1.30} (0.06) \\
\midrule
RT-1 \cite{rt1}    & ImageNet & 53.3\% & 22.2\% & 9.4\% & 3.8\% & 1.3\% & 0.90 \\
MT-R3M \cite{gr1} & Ego4D (R3M) & 52.9\% & 23.4\% & 10.5\% & 4.3\% & 1.8\% & 0.93 \\
RoboFlamingo \cite{flamingo} & OpenFlamingo 3B & 82.4\% & 61.9\% & 46.6\% & 33.1\% & 23.5\% & 2.48 \\
GR-1 \cite{gr1}    & Video Generative & \textbf{85.4\%} & \textbf{71.2\%} & \textbf{59.6\%} & \textbf{49.7\%} & \textbf{40.1\%} & \textbf{3.06} \\
\bottomrule
\end{tabular}
\end{table*}
\subsection{Results}
All \textit{DreamFormer} results are produced by us; baseline results are taken from published reports. Category-1 results are taken from HULC \cite{hulc}, except LUMOS, which we report from \cite{lumos}. Category-2 results are taken from GR-1 \cite{gr1}, except RoboFlamingo, which we report from \cite{flamingo}. Methods also differ in how much CALVIN supervision they use: \textit{DreamFormer} and all category-1 baselines train on the full play dataset, whereas the category-2 baselines train only on the language-annotated subset, roughly 1\% of the data. Note that the two axes act in opposite directions—category-2 methods benefit from large-scale external pretraining but use far less in-domain data—which is a further reason to compare within categories rather than across them.  Columns 1--5 report the success rate of completing that many instructions in a row, and average length is the mean number of instructions completed per chain. A chain terminates at the first failed instruction.

\textit{1) Single-Environment Comparison:} Table~\ref{tab:1} reports performance on the D→D split, the only setting for which LUMOS results are available and therefore the setting in which our contribution is most directly tested. We additionally include the remaining category-1 baselines. All results are averaged over 1000 instruction chains and three random seeds.

\textit{DreamFormer} outperforms LUMOS in all chain lengths, improving average length from 2.34 to 2.52, while operating on full-resolution observations rather than the $64{\times}64$ inputs LUMOS uses—a direct consequence of the scaled categorical VAE. While HULC remains the strongest method in this setting at 2.64 average length, it attributes much of its performance to performing control in the gripper-camera frame with a local policy. A variant of HULC without local   policy, in which the policy instead receives both camera views and acts in the base frame, reaches 2.24 average length, below \textit{DreamFormer}'s 2.52. Since \textit{DreamFormer}'s policy necessarily acts on a fused latent of both views, this variant is the closer point of comparison, though the two also differ in control frame. \textit{DreamFormer} is the best model-based agent reported on this setting and closes much of the gap to HULC, the best-performing baseline in this comparison, while learning its policy within imagination.
\begin{table*}[!t]
\caption{Ablation study on CALVIN D$\rightarrow$D split. The DreamFormer configuration
reported in Tables~\ref{tab:1} and~\ref{tab:2} is marked with $\dagger$.}
\label{tab:3}
\centering
\small
\begin{tabular}{lcccccc}
\toprule
\multirow{2}{*}{Configuration} 
 & \multicolumn{6}{c}{Number of Instructions in a Row (1000 chains)} \\
\cmidrule(lr){2-7}
 & 1 & 2 & 3 & 4 & 5 & Avg. Len. \\
\midrule
\rowcolor{gray!40} \multicolumn{7}{c}{\textit{pixel-shift augmentation strength (static / gripper)}} \\
\midrule
no shift
& 78.2\% (1.5) & 56.0\% (1.3) & 38.3\% (1.6) & 25.6\% (1.3) & 17.0\% (1.6) & 2.15 (0.07) \\
5 / 2 pixels
& 81.1\% (1.9) & 61.2\% (1.8) & 45.1\% (1.5) & 33.4\% (1.0) & 22.7\% (1.8) & 2.44 (0.08)  \\
10 / 4 pixels $\dagger$
& \textbf{82.3\% (1.7)} & \textbf{64.1\% (2.2)} & \textbf{47.1\% (2.8)} & \textbf{34.2\% (2.0)} & \textbf{24.3\% (2.0)} & \textbf{2.52 (0.11)}  \\
\midrule
\rowcolor{gray!40} \multicolumn{7}{c}{\textit{\textit{Component ablations at (5 / 2 pixels)}}} \\
\midrule
5 / 2 pixels (reference)
& 81.1\% (1.9) & 61.2\% (1.8) & 45.1\% (1.5) & 33.4\% (1.0) & 22.7\% (1.8) & 2.44 (0.08)  \\
w/o ResNet-18 encoder 
& 78.8\% (1.7) & 59.2\% (2.0) & 42.6\% (1.9) & 30.6\% (2.1) & 21.0\% (1.8) & 2.32 (0.09) \\
w/o latent scaling ($32\times32$)
& 75.9\% (1.9) & 53.4\% (1.2) & 35.6\% (0.1) & 22.6\% (0.4) & 14.0\% (1.0) & 2.02 (0.04) \\
w/o Dream Imitation 
& 70.3\% (2.2) & 45.2\% (1.0) & 27.5\% (2.0) & 17.5\% (1.4) & 9.9\% (1.0) & 1.70 (0.05)  \\
\bottomrule
\end{tabular}
\end{table*}

\textit{2) Multi-Environment and Zero-shot Comparisons:} Table~\ref{tab:2} reports performance
on the ABCD$\rightarrow$D and ABC$\rightarrow$D splits, covering multi-environment training and
zero-shot transfer to an environment unseen during training. \textit{DreamFormer}, HULC and MCIL
report results averaged over three random seeds, with standard deviations available for the
first two; the remaining baselines report single-run values.

On multi-environment training (ABCD$\rightarrow$D), \textit{DreamFormer} achieves comparable
performance to HULC among the methods without pretraining (3.11 $\pm$ 0.05 vs 3.06 $\pm$
0.07), a difference within the standard deviation of both.
Notably, \textit{DreamFormer} also exceeds RT-1 (2.45) and MT-R3M (2.08) despite their pretrained
visual backbones, indicating that pretraining alone does not determine performance in this
setting. GR-1 remains the strongest method overall at 4.21, which its own ablation attributes to
video generative pretraining \cite{gr1}.

On zero-shot transfer (ABC$\rightarrow$D), \textit{DreamFormer} reaches 1.30 average length,
nearly double HULC's 0.67, and again exceeds both RT-1 (0.90) and MT-R3M (0.93) despite their
pretrained visual backbones. Compared to HULC, the success rate of completing one to five tasks
in a row has increased by $20.5\%$, $18.9\%$, $12.6\%$, $7.5\%$, and $3.4\%$, respectively. Relative to its own
ABCD$\rightarrow$D performance, \textit{DreamFormer} retains 42\% of its average length in the
unseen environment while HULC retains 22\%. The zero-shot multi-environment setting demands that the agent solve manipulation tasks in a completely unfamiliar visual domain. Strong performance here indicates true semantic grounding—the ability to dynamically bind language concepts to novel visual observations, rather than relying on memorized spatial states. \textit{DreamFormer} improves substantially over HULC in this setting, suggesting that a task-agnostic world model trained on offline data captures dynamics that remain valid when the visual domain changes, whereas a policy cloned directly from demonstrations does not transfer as readily. GR-1 remains the strongest method overall at 3.06 average length.
\begin{figure*}[!t]
    \centering
    \includegraphics[width=\textwidth]{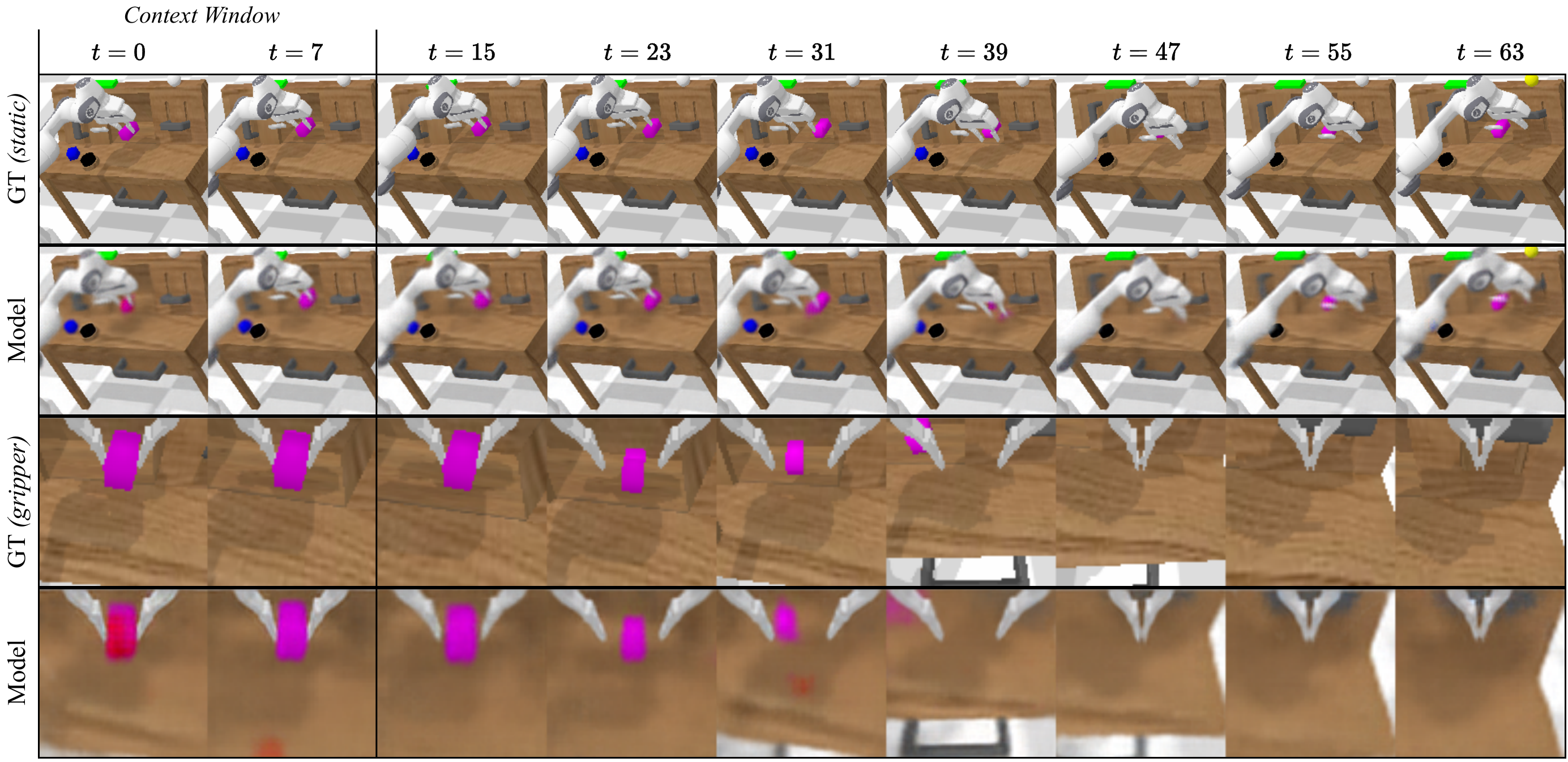}
    \caption{\textbf{Open-loop rollouts of the world model.} A trajectory from a held-out validation set. The
world model observes the first 8 frames ($t=0\ldots7$) as context and then autoregressively
predicts the remaining 56 latent states ($t=8\ldots63$), conditioned only on the ground-truth
action sequence. Rows show the ground truth (GT) and decoded
predictions (Model) for the static and gripper views; the vertical rule marks the end of the context
window, and frames are subsampled every eight steps. The arm pose and scene layout remain
accurate across the full horizon.}
    \label{fig:4}
\end{figure*}

\textit{3) Ablation Study:} Table~\ref{tab:3} reports the ablation study on the D→D split. We first vary the strength of the pixel-shift augmentation: no shift, 5/2 pixels, and 10/4 pixels for the static and gripper cameras respectively. Performance increases monotonically with shift magnitude at every chain length, from 2.15 to 2.44 to 2.52 average length, and we adopt the 10/4 setting in all other experiments. This is consistent with HULC \cite{hulc}, which reports a drop from 2.64 to 1.99 without image augmentation, though the effect is smaller in our setting. Note that HULC applies image augmentation during policy learning, whereas \textit{DreamFormer} applies it only during world model training. The remaining ablations remove individual components, each evaluated at the 5/2 setting. To determine whether the encoder architecture or the expanded latent capacity contributes more to our world model's performance, we evaluate two architectural ablations:
\begin{itemize}
    \item \textit{w/o ResNet-18 encoder:} replaces the ResNet-18 backbone and its spatial softmax layer with the shallow CNN encoder used in standard categorical VAEs \cite{dreamerv2}, while retaining the ($64\times32$) latent and the full training objective of Section~\ref{sec:4C}. This isolates the contribution of the encoder architecture.

    \item \textit{w/o latent scaling ($32\times32$):} reverts to the standard ($32\times32$) latent with the standard DreamerV2 objective—no entropy regularizer and fixed KL coefficients—while retaining the ResNet-18 encoder. Since the entropy regularizer and the KL schedule were introduced to stabilize the expanded latent and are not independently motivated at ($32\times32$), this variant tests the scaled latent as a package rather than isolating capacity from regularization.
\end{itemize}

Finally, to evaluate the efficacy of our policy learning framework, we evaluate the following ablation:
\begin{itemize}
    \item \textit{w/o Dream Imitation:} retains the imagined rollouts but replaces the intrinsic-reward optimization with behavioral cloning on the imagined trajectories, regressing the policy's actions onto the expert actions with an MSE loss.
\end{itemize}

Among the architectural components, the scaled latent contributes more than the encoder architecture. Replacing the ResNet-18 backbone with a shallow CNN costs 0.12 average length (2.44 to 2.32): at this resolution the shallow encoder still extracts usable features, and encoder capacity is not the binding constraint. Reverting to the standard ($32\times32$) latent with the standard objective costs 0.42 (2.44 to 2.02), and the gap widens with chain length—$14.0\%$ against $22.7\%$ at five instructions. The largest effect comes from the policy objective. Training the policy by behavioral cloning on the imagined rollouts, rather than by optimizing the intrinsic reward, drops performance to 1.70 average length, well below every architectural ablation. We attribute it to the distribution mismatch (covariate shift \cite{covariat-shift}) inherent in behavioral cloning.
\subsection{Imagination Rollouts Analysis}
We analyse imagination rollouts along two axes: the fidelity of the predicted states, and the computational cost of producing them. These are the two sides of the trade-off---reducing the tokens per observation lowers rollout cost, but only helps if the resulting latent still supports accurate long-horizon prediction. We first examine rollout fidelity qualitatively on held-out trajectories, then compare rollout cost against other Transformer world models using the configurations reported in their respective papers.

\textit{1) Rollout Fidelity:} To assess whether encoding a multi-view robotic observation into
a single input token retains enough information for long-horizon prediction, we perform open-loop
rollouts on held-out validation trajectories; Fig.~\ref{fig:4} shows an example. The world model
takes the first 8 frames as context and then autoregressively predicts the remaining 56 steps,
conditioned only on the ground-truth action sequence and receiving no further observations. The
predicted frames are decoded from \textit{the prior $\hat{z}$}, while the context frames are
decoded from \textit{the posterior $z$}.

Across the full horizon, the predicted arm pose and scene layout remain closely aligned with the
ground truth, and the manipulated block is tracked through the grasp, with its position drifting
slightly toward the end of the sequence. The gripper view is reconstructed less sharply than the
static view throughout, including within the context window; since those frames are decoded from
the posterior rather than predicted, this blur reflects the reconstruction fidelity of the
encoder--decoder rather than error accumulated during the rollout.

\begin{table*}[!t]
\caption{Sequence-model memory and attention cost for an imagination rollout of horizon $H$,
derived from the configuration reported in each paper. \textit{Predict steps/frame} is the
number of sequence-model passes needed to produce one imagined frame; \textit{Max cached tokens}
is the size of the key--value cache, which determines memory; \textit{Attention cost} counts query--key interactions over
the rollout, assuming key--value caching and matched model width and depth. $C$ is the length of
the fixed context window (3 frames for DINO-WM, 4 for CompACT). CompACT values are for its
navigation configuration.}
\label{tab:4}
\centering
\small
\begin{tabular}{llccc|cc}
\toprule
\multirow{3}{*}{Method} & \multirow{3}{*}{Tokenizer}
 & \multicolumn{3}{c|}{Design choices} & \multicolumn{2}{c}{Rollout cost ($H$ steps)} \\
\cmidrule(lr){3-5} \cmidrule(lr){6-7}
 & & Tokens & Predict & KV cache & Max cached & Attention \\
 & & /frame & steps/frame & type & tokens & cost \\
\midrule
DINO-WM \cite{dino-wm} & ViT patches (DINOv2)     & 196 & 1  & Fixed ($C$)  & $196\,C$ & $\mathcal{O}(196^2\, C H)$ \\
CompACT \cite{compact} & DINOv3 + resampler + FSQ & 8   & 4  & Fixed ($C$)  & $8\,C$   & $\mathcal{O}(4 \cdot 8^2\, C H)$ \\
IRIS \cite{iris}       & VQ-VAE                   & 16  & 16 & Growing ($H$) & $16H$    & $\mathcal{O}(16^2 H^2)$ \\
DreamFormer (ours) & Categorical VAE & 1 & 1 & Growing ($H$) & $H$ & $\mathcal{O}(H^2)$ \\
\bottomrule
\end{tabular}
\end{table*}

\textit{2) Rollout Cost:} We next examine how three design choices determine the cost of
imagination rollouts:
\begin{itemize}
    \item \textit{Tokenization:} the number of tokens each observation occupies, which sets the
    sequence length (\textit{Tokens/frame}).
    \item \textit{Next-frame prediction:} how the sequence model produces the tokens of the next
    frame---token by token, by iterative unmasking, or in a single pass
    (\textit{Predict steps/frame}).
    \item \textit{Context management:} whether the sequence model attends over the full rollout
    horizon or over a fixed window of recent frames (\textit{KV cache type}).
\end{itemize}
As discussed in Sec.~\ref{sec:3C}, the sequence length determines the memory footprint and the
quadratic cost of self-attention. Rather than timing each method's rollouts---which would
conflate these design choices with differences in model width, depth, and hardware---we derive the cost of a rollout of horizon $H$ directly from the configurations reported in each
paper. This isolates the effect of these three choices from implementation details.

Table~\ref{tab:4} compares \textit{DreamFormer} with three Transformer world models that differ along these three choices.
DINO-WM \cite{dino-wm} retains the full grid of frozen DINOv2 \cite{dinov2} patch features (196 tokens per frame) and uses a ViT-based predictor with frame-level causal attention to produce all tokens of the next frame in a single pass, attending over a fixed window of 
$C$ past frames. CompACT \cite{compact} resamples frozen DINOv3 \cite{dinov3} features into 8 discrete tokens through learnable queries and finite scalar quantization (FSQ); its navigation model produces the next frame's tokens by masked generative decoding over 4 unmasking steps, again conditioned on a fixed window of $C$ past frames. IRIS \cite{iris} encodes each frame into 16 discrete tokens with a VQ-VAE \cite{discreteAE}, and a GPT-style Transformer predicts them one at a time over a growing cache of the full rollout. \textit{DreamFormer}, by contrast, encodes each frame into a single token, predicts it in one pass, and likewise retains the full rollout in a growing cache.

We derive the rollout cost, reported in the last two columns of Table~\ref{tab:4}, from these three choices, as follows:
\begin{itemize}
    \item \textit{Max cached tokens} measures the memory footprint of the sequence model: the largest number of tokens held in the key--value cache during the rollout, equal to \textit{Tokens/frame}$\times H$ for a growing cache and \textit{Tokens/frame}$\times C$ for a fixed window, independent of $H$.

    \item \textit{Attention cost} measures computation: the total number of query--key interactions over the rollout. With a growing cache every token attends over all preceding tokens, giving a cost quadratic in both the tokens per frame and $H$; with a fixed window each token attends only over the window, giving a cost linear in $H$. Masked generative prediction re-queries all tokens of a frame at each unmasking pass, which contributes CompACT's additional factor of 4.

\end{itemize}

A fixed window thus trades memory for context: it bounds the cache and makes attention cost linear in $H$, but discards every state outside the fixed window, so that once the rollout exceeds $C$ steps no real observation remains in context and the model conditions only on its own predictions. Such bounded context suits short-horizon planning, the setting in which DINO-WM and CompACT's navigation model are applied through MPC, but limits long-horizon imagination. Linear scaling in $H$ does not, however, make these methods cheaper. As Table~\ref{tab:4} shows, their cost carries the square of the tokens per frame ($196^2$ for DINO-WM and $4 \cdot 8^2$ for CompACT), a factor that exceeds the horizon itself at any practical rollout length, including our $H=32$. IRIS retains the full rollout, but its 16 tokens per frame multiply its quadratic cost by $16^2$ relative to \textit{DreamFormer}. By representing each observation with a single token, \textit{DreamFormer} is the only method among those compared that retains the full rollout in context at the lowest attention cost.
\section{Implementation and Training Details}
\subsection{World Model}
The world model consists of the scaled categorical VAE and the Transformer-based sequence model. The static and gripper views are processed by two independent ResNet-18 encoders trained from scratch, each followed by a spatial softmax layer \cite{spatialsoftmax}. Their outputs are concatenated and passed through a fully connected layer to obtain visual features, which a second fully connected layer maps to the logits of the posterior: a categorical distribution over 64 categories of 32 classes each. The sampled latent $z_t$ is flattened into a 2048-dimensional binary vector with 64 active entries, and gradients are propagated through the sample by straight-through estimation \cite{straight-through}. At each time step, $z_t$ and the action $a_t$ are concatenated and mapped by a two-layer MLP to a single input token, so that each time step occupies one position in the sequence. Reconstruction is handled by two separate transposed-convolutional decoders, one per view, each taking the flattened latent $z_t$ as its only input and upsampling to the native resolution of that view ($200\times200$ for the static camera, $84\times84$ for the gripper camera). Both decoders are used only for the reconstruction objective and for visualization, and are not involved in imagination or policy learning.

The sequence model is a GPT-style Transformer \cite{gpt} with causal attention, whose output $h_t$ is mapped by a fully connected layer to the logits of the prior from which the next latent state $\hat{z}_{t+1}$ is sampled.
The world model has 60M parameters and was trained using a distributed setup of four NVIDIA Quadro RTX 6000 GPUs. The training procedure utilized a sequence length of 64 and a per-GPU batch size of 4, resulting in an effective batch size of 16. All architectural and optimization hyperparameters are detailed in Table \ref{tab:5}.
\begin{table}[t]
\caption{World model hyperparameters.}
\label{tab:5}
\centering
\begin{tabular}{l c c}
\toprule
\textbf{Parameter} & \textbf{Symbol} & \textbf{Value} \\
\midrule
Batch size                      & $B$       & 16 \\
Sequence length                 & $T$       & 64 \\
Transformer blocks              & --        & 5 \\
Transformer heads               & --        & 12 \\
Transformer embedding dim.      & --        & 768 \\
Stochastic latent state         & $z_t$     & $64 \times 32$ (2048) \\
Model latent state              & $s_t$     & 2816 \\
Optimizer                       & --        & AdamW \\
Learning rate                   & --        & \num{1e-4} \\
Adam epsilon                    & $\epsilon$ & \num{1e-5} \\
Weight decay                    & --        & \num{5e-2} \\
Gradient clipping               & --        & 100 \\
Reconstruction loss coef.       & $\beta_1$ & 0.05 \\
Entropy regularizer coef.       & $\beta_2$ & 10 \\
Representation KL coef.         & $\beta_3$ & $0.1 \rightarrow 0.5$ \\
Dynamics KL coef.               & $\beta_4$ & $1.0 \rightarrow 2.0$ \\
KL schedule                     & --        & cosine, first 250K steps \\
Pixel-shift augmentation        & --        & 10 / 4 (static / gripper) \\
\bottomrule
\end{tabular}
\end{table}

\subsection{Actor--Critic Agent}
We follow LUMOS \cite{lumos} for the architecture and optimization of the actor--critic agent, so that differences in performance against it are attributable to the world model rather than to the policy learning setup. The only change is the dimension of the model state $s_t = \{h_t, z_t\}$: because \textit{DreamFormer} expands the categorical latent to $(64\times32)$, its model state is 2816-dimensional against LUMOS's 2048. The actor--critic agent has 47M parameters and was trained using a single GPU per seed. All architectural and optimization hyperparameters are detailed in Table \ref{tab:6}.

\begin{table}[t]
\caption{Actor--Critic hyperparameters.}
\label{tab:6}
\centering
\begin{tabular}{l c c}
\toprule
\textbf{Parameter} & \textbf{Symbol} & \textbf{Value} \\
\midrule
Batch size                      & $B$       & 512 \\
Imagination horizon             & $H$       & 32 \\
Context window                  & $\mathcal{C}$  & 16 \\
Latent goal dimensions          & $g$         & 32 \\
Latent plan categories          & --         & 32 \\
Latent plan classes             & --        & 32 \\
KL loss scale                   & $\alpha_1$     & 0.1 \\
Contrastive loss scale          & $\alpha_2$     & 3.0 \\
Discount factor                 & $\gamma$       & 0.995 \\
TD($\lambda$) parameter         & $\lambda$      & 0.95 \\
Optimizer                       & --        & Adam \\
Actor learning rate             & --        & \num{2e-4} \\
Critic learning rate            & --        & \num{3e-4} \\
Target critic update interval   & --        & 100 \\
\bottomrule
\end{tabular}
\end{table}

\section{Conclusion and future work}
We presented \textit{DreamFormer}, a model-based robotic agent that learns a task-agnostic
Transformer world model from unstructured play data and acquires language-conditioned,
multi-task behaviors entirely within its latent imagination. To make long-horizon imagination
affordable at full resolution, we scaled the categorical VAE with a ResNet-18 backbone and an
expanded categorical latent space, encoding multi-view observations at a single token per
timestep. Sequence length therefore grows only with the imagination horizon and is independent
of input resolution, keeping the memory and attention cost of a rollout bounded by the horizon
alone.

On the CALVIN benchmark, \textit{DreamFormer} outperforms LUMOS, the comparable model-based
agent, on single-environment evaluation, and nearly doubles the performance of the behavior cloning baseline on zero-shot transfer to an unseen environment. Notably, this level of transfer is reached with a visual representation learned from CALVIN play data alone, without initialization from externally pretrained models, suggesting that what transfers is the dynamics captured by the world model rather than the breadth of the data its encoder was trained on. Our ablations show that the scaled latent contributes more than the encoder architecture, and that the largest single factor is the policy objective: replacing the intrinsic reward with behavior cloning on the same imagined rollouts costs more than any architectural component.

Several points remain for future work. Our evaluation is confined to a simulated benchmark, and
whether the same representation transfers to real robots is untested. Encoding an entire multi-view
observation at one token bounds how much spatial detail the latent can carry, which our
qualitative rollouts suggest is felt most in small manipulated objects rather than in the arm or
scene layout. Extending the approach to real-robot data, and to policies that consume
gripper-centric detail at higher fidelity than the shared latent provides, are our next steps for future work.

\section{Acknowledgment}
Mostafa Kotb is funded by a scholarship from the Ministry of Higher Education of the Arab Republic of Egypt.

\bibliographystyle{IEEEtran}
\bibliography{IEEEabrv,references.bib}

\end{document}